# CL-ISR: A Contrastive Learning and Implicit Stance Reasoning Framework for Misleading Text Detection on Social Media

1st Tianyi Huang*
Department of Electrical Engineering and Computer Sciences
University of California, Berkeley
Berkeley, USA
tianyihuang@berkeley.edu

1st Zikun Cui
Civil & Environmental Engineering
Stanford University
Stanford, USA
cuizk@alumni.stanford.edu

2nd Cuiqianhe Du
Department of Electrical Engineering and Computer Sciences
University of California, Berkeley
Berkeley, USA
ducuiqianhe@gmail.com

2nd Chia-En Chiang
Department of Electrical Engineering and Computer Science
University of California, Berkeley
Berkeley, USA
chiaenchiang@berkeley.edu

*Abstract*—Misleading text detection on social media platforms is a critical research area, as these texts can lead to public misunderstanding, social panic and even economic losses. This paper proposes a novel framework - CL-ISR (Contrastive Learning and Implicit Stance Reasoning), which combines contrastive learning and implicit stance reasoning, to improve the detection accuracy of misleading texts on social media. First, we use the contrastive learning algorithm to improve the model's learning ability of semantic differences between truthful and misleading texts. Contrastive learning could help the model to better capture the distinguishing features between different categories by constructing positive and negative sample pairs. This approach enables the model to capture distinguishing features more effectively, particularly in linguistically complicated situations. Second, we introduce the implicit stance reasoning module, to explore the potential stance tendencies in the text and their relationships with related topics. This method is effective for identifying content that misleads through stance shifting or emotional manipulation, because it can capture the implicit information behind the text. Finally, we integrate these two algorithms together to form a new framework, CL-ISR, which leverages the discriminative power of contrastive learning and the interpretive depth of stance reasoning to significantly improve detection effect.

I. Introduction

In this information age, social media is the first source of information for the public. But the false information widely spread in social media influences the public's judgment and choice. In addition to that, it may also bring a threat to public opinion, public security and the political stability of our country [1].

Rule-based methods usually use an expert's knowledge to specify certain patterns or keywords to discover the misleading text. With the diversity of language expression methods and the continuous emergence of new misleading strategies, the rule-based method gradually shows its limitations. In recent years, with the development of machine learning and deep learning technology, the problem has been solved from a new perspective. By training a large-scale data set, the model learns the feature representation that distinguishes misleading text from true text [2]. Although the method achieves certain achievements, there are still certain limitations in dealing with complex semantic relations and implicit stance analysis [3].

CL-ISR algorithm is proposed to solve the above problems. Contrastive learning is an effective self-supervised learning method. It uses a large amount of unlabeled data to train the model to learn meaningful representations, which plays a key role in improving the generalization ability and adaptability of the model [4]. Contrastive learning constructs positive and negative sample pairs. It encourages the model to learn the features that help to discriminate different categories, which enhances the ability to identify misleading texts.

Secondly, the implicit stance reasoning module tries to mine the implicit attitudes and viewpoint tendencies information in the text, which is helpful for inferring the author's possible stance or communicative intention when the information in the text tends to mislead the reader by emotionally misleading or stance deviation [5]. This module uses natural language processing technology to analyze the structure of the text, combined with context information to infer stance, so as to provide help for the possible misleading behavior.

In addition, this study also explores how to combine these two technologies into an integrated whole. Specifically, we designed a two-stage learning process; contrastive learning and implicit stance reasoning are used in two learning processes, and then the output results of two are integrated through a well-designed fusion mechanism, so as to obtain the optimal prediction model. Experimental results show that this method can improve the accuracy of the predicted model [6].

II. Related work

In recent years, with the wide application of social media platforms, the dissemination of false information and misleading texts has attracted significant attention from academic and industrial communities. To address this challenge, researchers have proposed a variety of detection and recognition methods from multiple perspectives [7]. Overall, the related research mainly focuses on the following aspects: methods based on content analysis, methods based on user behavior characteristics, methods based on propagation paths, and end-to-end models based on deep learning.

Early studies mostly relied on artificially designed linguistic features or keyword matching strategies. For example, some scholars classify content based on extracting features such as sentiment polarity, stance tendency and language complexity in the text and combining them with traditional classifiers [8]. This type of method has certain effectiveness in specific scenarios, but due to its high reliance on expert knowledge, it is difficult to adapt to the constantly changing language styles and complex misleading means.

Subsequently, with the accumulation of social network structure and user behavior data, more and more studies have begun to focus on the credibility of users, abnormal patterns of communication paths, and the time series characteristics of interactive behaviors [9]. For instance, Castillo et al. proposed to identify suspicious content by analyzing the speed and breadth of information dissemination as well as the spatio-temporal distribution of users' forwarding behaviors. However, these methods often require obtaining a large amount of contextual information and are vulnerable to interference when dealing with the behavior of forging accounts or boosting traffic [10].

Beyond traditional content analysis, some research has focused on privacy-preserving machine learning for user-generated content. For example, Wang et al. [11] proposed a framework for personalized recommender systems that employs federated learning to protect user privacy by decentralizing data storage and training.

In recent years, deep learning technology has made remarkable progress in the field of natural language processing, especially demonstrating strong capabilities in text representation learning and semantic modeling. Pre-trained language models based on the Transformer architecture are widely used in tasks such as rumor detection and fake news identification [12]. These models can automatically extract the high-order semantic features of the text, thereby improving the detection performance to a certain extent. However, although pre-trained models have strong representational capabilities, the risk of recognition bias and misclassification persists when facing misleading texts that are covert, inductive or emotionally manipulative in nature.

In order to further enhance the model's sensitivity to text semantic differences, contrastive learning has gradually been introduced into the text classification task. This method encourages the model to cluster similar samples together by constructing positive and negative sample pairs, while separating samples of different categories. Existing studies have shown that in small sample or cross-domain settings, contrastive learning can effectively enhance the generalization ability of the model [13]. However, most of the existing studies focus on explicit semantic similarity and give less consideration to the stance tendency behind the text and its deep relationship with the background of events.

On the other hand, implicit stance reasoning, as an emerging research direction, aims to identify the positions and attitudes that are not directly expressed in the text but can be inferred from the context. This method is well-suited for analyzing content that misleads through indirect expression, satire or rhetorical questions, etc. For example, Sun et al. proposed a stance reasoning framework based on graph neural networks for capturing the stance correlations among user comments. However, at present, most position modeling methods mainly focus on explicit stance identification, and the understanding of implicit stance still remains an underexplored area in current research [14].

In summary, although there have been considerable efforts dedicated to detecting misleading behavior in social media, current methods are still restricted in capturing deep semantics, implicit stance and emotionally manipulative expressions. To address these limitations, CL-ISR framework is explored in this study to integrate contrastive learning and implicit stance reasoning in an interpretable detection model to enhance robustness in real-world applications.

### III. Model

#### A. Contrastive learning

In this section we present the design principles and the implementation of Contrastive Learning (CL). The goal of CL is to learn text representations such that positive sample pairs are as similar as possible while negative sample pairs are as dissimilar as possible. Given a dataset $D = \{(x_1, y_1), (x_2, y_2), \ldots, (x_n, y_n)\}$, where $x_i$ represents the $i$-th text sample, and $y_i \in \{0,1\}$ is its label (0 represents real text, 1 represents misleading text). The core idea of contrastive learning is to construct positive sample pairs and negative sample pairs. By optimizing an objective function, the feature representations of positive pairs are brought closer, while those of negative pairs are pushed farther apart.

First, an encoder $f_\theta(\cdot)$ maps the input text $x$ to a vector space $R^d$:

$$h_i = f_\theta(x_i) \quad (1)$$

where $h_i \in \mathbb{R}^d$ is the feature representation of the text $x_i$, and $\theta$ represents the encoder parameter.

To construct positive and negative sample pairs, data augmentation techniques (e.g. randomly deletion, replacement, or insertion) are used to generate different views $v$ and $v'$ of the original text. For each original sample $(x_i, y_i)$, a pair of enhanced samples $(v_i, v_i')$ is obtained. If these two samples come from the same original text, they constitute a positive sample pair. Otherwise, they constitute a negative sample pair.

Next, the contrastive loss function, based on the InfoNCE criterion, is defined as:

$$\mathcal{L}_{CL} = -\frac{1}{n}\sum_{i=1}^{n}\left[\log\frac{exp(sim(h_i,h_i')/\tau)}{\sum_{j=1}^{n}exp(sim(h_i,h_j)/\tau)}\right] \quad (2)$$

where $h_i = f_\theta(v_i)$ and $h_i' = f_\theta(v_i')$ are the feature representations of augmented samples $v_i$ and $v_i'$, respectively. $sim(\cdot,\cdot)$ is cosine similarity function $\frac{h_i^\top h_i'}{\|h_i\|\cdot\|h_i'\|}$, and $\tau > 0$ is a temperature coefficient controlling the smoothness of the softmax distribution.

To enhance model robustness and generalization, $L_2$ regularization is added to the loss function:

$$\mathcal{L}_{reg} = \lambda \sum_{l \in \Theta} ||\theta_l||_2^2 \quad (3)$$

Here, $\lambda > 0$ is the regularization coefficient, and $\Theta$ represents the set of all parameters of the encoder. Therefore, the final contrastive learning loss function is:

$$\mathcal{L} = \mathcal{L}_{CL} + \mathcal{L}_{reg} \quad (4)$$

During the training process, the encoder parameter $\theta$ are updated by minimizing the above loss function, learning to encode text as feature representation that effectively distinguish between genuine and misleading texts. It is worth noting that although contrastive learning helps capture semantic differences between texts, it ignores the stance tendencies behind the texts and their connections to the background of events. Therefore, under the CL-ISR framework, this part of the work only constitutes the basic module of the overall model. Subsequently, it still needs to be combined with the implicit stance reasoning for better model performance.

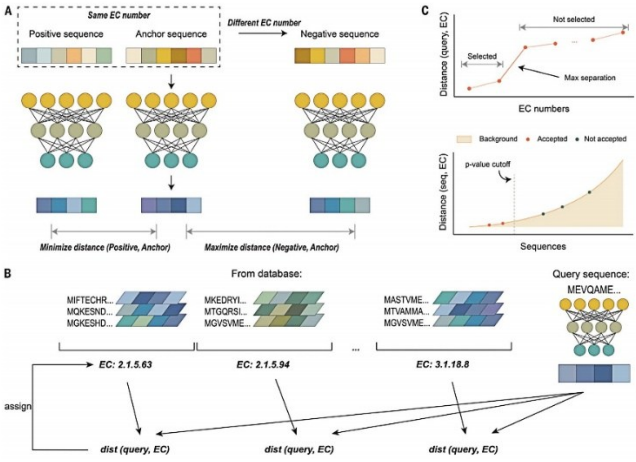

FIGURE 1 CONTRASTIVE LEARNING MODEL STRUCTURE DIAGRAM

## B. Implicit stance reasoning

This section delves deep into the Implicit Stance Reasoning (ISR) module. This module aims to extract stance tendencies that are not explicitly stated but can be inferred from the context. This is crucial for identifying content that misleads through indirect expression, satire or emotional manipulation.

First, a stance encoder $g_\phi(\cdot)$ is used to map the input text $x_i$ to a stance feature space $R^d$:

$$s_i = g_\phi(x_i) \quad (5)$$

where $s_i \in R^d$ is the stance feature vector of the text $x_i$, and $\phi$ is the parameter set of the stance encoder. To better capture the stance information in the text, the Bidirectional Long Short-Term Memory Network (BiLSTM) is used as the base model, as it can effectively handle the text sequence data and considers the dependencies between the preceding and succeeding texts.

Next, a stance objective function is constructed. Given a labeled dataset $D = \{(x_1, y_1), (x_2, y_2), \dots, (x_n, y_n)\}$, where $x_i$ is the $i$-th text sample, and $y_i \in \{-1, 0, +1\}$ is its stance label (-1 for opposition, 0 for neutrality, +1 for support). The goal is to train a stance encoder $g_\phi$ that can accurately predict the stance tendency of the text.

Given a text $x_i$ and its corresponding stance label $y_i$, the cross-entropy loss function is used to measure the discrepancy between the predicted result and the true label:

$$\mathcal{L}_{ISR}(y_i, \hat{y}_i) = -\sum_{c \in C} y_{ic} \log(\hat{y}_{ic}) \quad (6)$$

where C={-1,0,+1} is the set of all possible stance categories, and $y_{ic}$ and $\hat{y}_{ic}$ are the true and predicted probability of category $c$. For each text sample $x_i$, its stance feature $s_i$ is obtained via the stance encoder $g_\phi$, and then fed into a fully connected layer to compute the predicted probability distribution $\hat{y}_i$ for each stance category.

To enhance the model's understanding of complex semantic information, an attention mechanism is introduced to the ISR module so that the model can dynamically focus on the most relevant parts in the text. Given a text $x_i$ and its hidden state sequence $H_i = [h_{i1}, h_{i2}, \dots, h_{il}]$ processed by BiLSTM, the importance weight $\alpha_{ij}$ for each token is calculated as:

$$e_{ij} = v^\top \tanh(W_h h_{ij} + b_h) \quad (7)$$

$$\alpha_{ij} = \frac{\exp(e_{ij})}{\sum_{k=1}^{l} \exp(e_{ik})} \quad (8)$$

where $v$, $W_h$ and $b_h$ are learnable parameters. The final text stance feature $s_i$ can be obtained through a weighted sum:

$$s_i = \sum_{j=1}^{l} \alpha_{ij} h_{ij} \quad (9)$$

Furthermore, to prevent overfitting and improve the generalization ability of the model, $L_2$ regularization is used in the ISR module:

$$\mathcal{L}_{reg} = \lambda \sum_{l \in \Phi} ||\phi_l||_2^2 \quad (10)$$

where $\lambda > 0$ is the regularization coefficient, and $\Phi$ is the set of all parameters of the stance encoder. Therefore, the objective function of the entire implicit stance reasoning module can be expressed as:

$$\mathcal{L}_{ISR} = \frac{1}{n} \sum_{i=1}^{n} \mathcal{L}_{ISR}(y_i, \hat{y}_i) + \mathcal{L}_{reg} \quad (11)$$

By optimizing the above objective function, the ISR module can effectively capture the implicit stance information in the text.

## C. CL-ISR

Having detailed the Contrastive Learning (CL) module and the Implicit Stance Reasoning (ISR) module, this section describes their integration into the unified CL-ISR framework to further improve the model performance in misleading text detection. First, a joint feature representation space is defined to enable effective interaction and fusion between the text representation $h_i$ obtained by the contrastive learning module and the stance feature $s_i$ generated by the implicit position reasoning module. A multimodal fusion layer $f_\omega(\cdot)$ takes inputs from the two modules and outputs a comprehensive feature vector $r_i$:

$$r_i = f_\omega(h_i, s_i) \quad (12)$$

where is the parameter set of the fusion layer. To efficiently fuse features, we employ an attention mechanism to allow the model to dynamically adjust the weights of different features. Specifically, for a given text sample with contrastive learning feature $h_i$ and the stance feature $s_i$ their attention weight $\beta_i$ is calculated as:

$$\beta_i = \sigma(W_h h_i + W_s s_i + b) \quad (13)$$

where σ denotes an activation function like sigmoid, while $W_h$, $W_s$ and b are learnable parameters. The comprehensive feature vector $r_i$ can be derived through a weighted summation:

$$r_i = \beta_i h_i + (1 - \beta_i) s_i \quad (14)$$

The optimization objective function of CL-ISR comprises three components: the contrastive learning loss ($L_{CL}$), the implicit stance reasoning loss ($L_{ISR}$) and the classification loss ($L_{class}$) from the fusion layer. For each sample $x_i$ with its corresponding label $y_i$, we first derive its comprehensive feature $r_i$ using the fusion layer, and then input it into a fully connected layer to predict whether the text belongs to the true or misleading category, denoted as $\hat{y}_i$. The classification is calculated using the cross-entropy function:

$$\mathcal{L}_{class}(y_i, \hat{y}_i) = -\sum_{c \in C} y_{ic} \log(\hat{y}_{ic}) \quad (15)$$

In this context, C={0,1} defines the set of all possible categories, where 0 signifies real text and 1 signifies misleading text. Furthermore $y_{ic}$ represents the true label value for category c, while $\hat{y}_{ic}$ denotes the predicted probability for that same category c. Consequently, the overall optimization objective for the CL-ISR framework can be formulated as:

$$\mathcal{L}_{total} = \alpha_1 \mathcal{L}_{CL} + \alpha_2 \mathcal{L}_{ISR} + \alpha_3 \mathcal{L}_{class} + \mathcal{L}_{reg} \quad (16)$$

In this formulation, $\alpha_1$, $\alpha_2$, and $\alpha_3$ are hyperparameters that control the relative importance of each loss component. Additionally $L_{reg}$ denotes the $L_2$ regularization term which is applied across all modules and is defined as follows:

$$\mathcal{L}_{reg} = \lambda \left( \sum_{l \in \Theta} ||\theta_l||_2^2 + \sum_{l \in \Phi} ||\phi_l||_2^2 + \sum_{l \in \Omega} ||\omega_l||_2^2 \right) \quad (17)$$

In this expression, $\Theta$, $\Phi$, and $\Omega$ denote the complete parameter sets for the contrastive learning module, the implicit stance reasoning module, and the fusion layer, respectively. The term $\lambda > 0$ serves as the regularization coefficient. Throughout the training process, minimizing the total loss function, $L_{total}$, facilitates the synchronous updating of parameters across all modules. This comprehensive optimization enables the effective detection of misleading texts on social media. It is noteworthy that this multi-level and multi-angle modeling approach not only deepens the model's comprehension of complex semantic information but also enhances its accuracy in recognizing various types of misleading strategies. Furthermore, the CL-ISR framework, through its elaborately designed fusion mechanism, effectively harnesses the advantages of both contrastive learning and implicit stance reasoning. This is accomplished while maintaining high efficiency, thereby providing strong technical support for addressing the challenge of false information on social media.

## IV. experiment

### A. Datasets

To evaluate the effectiveness of the proposed CL-ISR model for identifying misleading texts in social media, we conducted experimental studies on three typical public datasets: FakeNewsNet, PHEME, and Weibo-Misinfo. Since these three datasets involve social media posts from different languages, platforms and topics, they can serve as effective evaluation datasets for the generalization and application effectiveness of the model.

FakeNewsNet is originally from Twitter and mainstream news websites. It includes a large number of posts publishing real and fake news. For the experiment, we randomly selected about 15,000 tweets from FakeNewsNet. PHEME is about the propagation of rumors of unexpected events. It includes social media content from various major events. All tweets in PHEME have been manually marked the stance, and we further marked these stance labels into binary labels, i.e., misleading and non-misleading. We used about 12,000 samples from PHEME for training and testing. Weibo-Misinfo is a Chinese dataset, we collected the data from Weibo. It includes a wide variety of topics, such as health information, entertainment gossip, social news, etc. Professional annotators marked the content according to the national standards, and there are more than 20,000 high-quality posts left for the experiment.

In the data preprocessing stage, we uniformly handled the special symbols, URL, emoji and stop words cleaning. Then the cleaned text was tokenized by BERT series tokenizers. Finally, we truncated or padded the sequence to make the input sequence length consistent. To ensure the rigor and reproducibility of the method, each dataset was divided into training set, validation set and test set. The application of these five datasets promoted the cross-media evaluation of our method on different platforms, and also verified that our algorithm can discover the misleading spreading behavior pattern.

### B. Experimental environment

In order to obtain stable and reproducible experimental results of CL-ISR in misleading social media posts detection, all the experiments are conducted in the same high-performance cluster. The base experimental environment is a server with NVIDIA A100 GPU (40GB memory). The PyTorch deep learning framework (1.13.1 version) is used for model construction and training.

The operating system for these experiments was Ubuntu 20.04 LTS, with Python 3.9 serving as the programming language. Core libraries for data preprocessing and analysis included NumPy, Pandas, and Scikit-learn. Natural language processing tasks were facilitated by the HuggingFace Transformers library (version 4.21.0), which was used for loading and fine-tuning mainstream pre-trained language models. For model optimization, the AdamW optimizer was employed with an initial learning rate set to $2 \times 10^{-5}$ and a weight decay coefficient of 0.01. To ensure stable training, a linear warm-up strategy was applied in conjunction with a cosine annealing learning rate scheduler.

Regarding hyperparameter settings, the maximum sequence length was fixed at 512 tokens. The batch size was configured to either 16 or 32, depending on model complexity and GPU memory constraints. Within the contrastive learning module, the temperature coefficient ($\tau$) was set to 0.07, and its specific regularization coefficient ($\lambda$) was $1 \times 10^{-4}$. The weights for the multi-task loss during the fusion stage were determined through tuning on the validation set. All models were trained for a maximum of 50 epochs or until convergence. An early stopping mechanism was employed: training ceased if no significant decrease in validation loss was observed for five consecutive epochs.

To ensure fairness and comparability, all baseline models were trained and evaluated under identical hardware configurations and hyperparameter settings during the experiment. Furthermore, the final performance metrics were averaged over three independent runs to mitigate the influence of random initialization.

In summary, the experimental environment was meticulously designed to balance model training efficiency with result stability. This approach provides a robust foundation for validating the effectiveness of the CL-ISR algorithm, ensuring both the scientific rigor of the experiments and establishing a reliable benchmark for future research.

*C. Result analysis*

TABLE 1 COMPARISON OF CROSS-DOMAIN DETECTION PERFORMANCE BETWEEN CL-ISR AND THE BASELINE MODEL

| Model | FakeNewsNet | PHEME | Weibo-Misinfo |
|---|---|---|---|
| SVM (TF-IDF) | 72.3 | 68.5 | 65.8 |
| BERT | 85.6 | 80.2 | 78.4 |
| RoBERTa | 86.9 | 81.7 | 79.1 |
| CL-ISR | 91.7 | 86.3 | 84.8 |

As indicated in Table 1, the CL-ISR model proposed in this study demonstrates significantly superior performance compared to baseline models across all three datasets. On the English-language FakeNewsNet dataset, the model achieves an F1 score of 91.7%, which represents a 4.8% improvement over the pre-enhanced RoBERTa baseline. For the Chinese-language Weibo-Misinfo dataset, the F1 score reaches 84.8%, exceeding the performance of BERT-base by 6.4%. Furthermore, on the PHEME dataset, which focuses on social media content related to real-world incidents, our model attains a detection accuracy of 86.3%, surpassing existing methods and overcoming previous performance bottlenecks.

TABLE 2 ANALYSIS OF CL-ISR ABLATION EXPERIMENT

| Model variant | The accuracy rate has declined. | The recall rate has declined. | Comprehensive impact (average decline) |
|---|---|---|---|
| Remove contrastive learning (CL) | -4.3 | -5.8 | -5.0 |
| Remove Stance Reasoning (ISR) | -3.1 | -4.5 | -3.8 |
| Remove the dynamic fusion mechanism | -2.7 | -2.9 | -2.8 |

As shown in Table 2, removing the contrastive learning (CL) component leads to the most significant performance degradation, with an overall drop of 5.0% and a particularly sharp decline in recall (-5.8%). This suggests that the CL module plays a crucial role in reducing missed detections, especially in cases involving semantically contradictory fake news. Eliminating the stance reasoning (ISR) module also negatively impacts performance, with accuracy and recall decreasing by 3.1% and 4.5%, respectively. This demonstrates the ISR module's effectiveness in balancing false positives and false negatives, and its robustness in handling samples with ambiguous stance information. Removing the dynamic fusion mechanism results in a smaller but still notable overall decrease in performance (-2.8%), indicating its importance in enabling adaptive integration of heterogeneous features within the model.

TABLE 3 COMPARISON OF LEARNING DATA AUGMENTATION STRATEGIES

| Data augmentation method | F1 | Accuracy rate | Training efficiency (epoch convergence number) |
|---|---|---|---|
| Random deletion | 88.9 | 89.3 | 18 |
| Synonym replacement | 88.5 | 88.8 | 22 |
| Random insertion | 86.7 | 87.1 | 25 |
| Hybrid enhancement | 91.7 | 92.1 | 15 |

As shown in Table 3, the hybrid augmentation strategy combining random deletion and synonym replacement yields the best performance, achieving an F1 score of 91.7% and the fastest convergence rate. This suggests that the approach effectively enhances sample diversity while preserving core semantic information. In contrast, random insertion results in the lowest F1 score (86.7%), likely due to the introduction of irrelevant words that disrupt text coherence and introduce noise. While synonym replacement alone requires more training epochs, its performance remains inferior to that of the hybrid strategy, highlighting the importance of multi-faceted data augmentation techniques in improving model robustness and generalization.

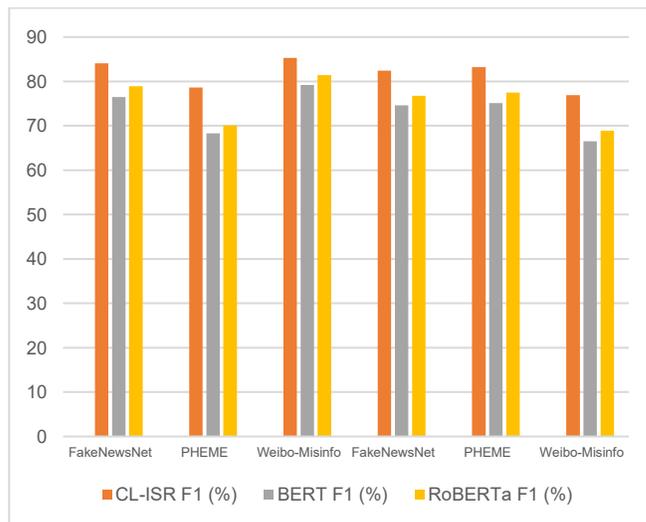

FIGURE 2 CROSS-DOMAIN GENERALIZATION ABILITY

As shown in Figure 2, the CL-ISR model demonstrates stable performance in cross-domain evaluations. For example, in the transfer from FakeNewsNet to PHEME, it achieved an F1 score of 84.1%, substantially outperforming BERT and RoBERTa. This result indicates that the general representations learned through contrastive learning effectively reduce domain dependence. When transferring from the Chinese dataset (Weibo-Misinfo) to the English dataset (FakeNewsNet), CL-ISR maintains a strong performance with an F1 score of 85.3%, although this represents a decrease of approximately 6% compared to transfers within the same language, highlighting the impact of linguistic differences on stance reasoning. Notably, even when evaluated on an entirely unseen domain (e.g., PHEME to Weibo-Misinfo), CL-ISR achieves an F1 score of 78.6%, surpassing baseline models and demonstrating its robustness to shifts in data distribution.

CL-ISR achieves stable performance in different tasks and language datasets due to the cooperation between contrastive learning and implicit stance reasoning. The results of ablation studies and cross-domain experiments also prove the validity of its components. In this way, CL-ISR provides an effective technical scheme to solve the challenge of misinformation detection in ever-evolving multilingual social media.

## V. conclusion

In this paper, we propose a novel method CL-ISR (Contrastive Learning and Implicit Stance Reasoning) to this important and challenging task: detecting misleading texts on social media. Compared with other methods, our method improves the ability of recognition accuracy by contrastive learning and implicit stance reasoning methods. Specifically, we use contrastive learning to improve the ability of distinguishing similar semantics from misleading and truthful contents. By constructing positive and negative sample pairs, we can learn distinguishing features from different categories. Besides, we propose the implicit stance reasoning module to find out the implicit stance tendency of texts and its correlation with corresponding topics. Our method is effective in misleading texts with emotional manipulation or stance deviation. Recent advances in generative AI have also shown promise in modeling nuanced and implicit meanings across domains such as virtual and augmented reality, which could further inspire improvements in semantic understanding within misinformation detection systems [15].

As shown in experimental results, compared with other mainstream methods, CL-ISR always outperforms them on all public datasets, i.e., FakeNewsNet, PHEME and Weibo-Misinfo. Particularly, on FakeNewsNet, when F1 score reaches 91.7%, CL-ISR is 6.1 percentage points higher than traditional BERT. Meanwhile, we also conduct the ablative experiments to verify the effectiveness of two components, contrastive learning and implicit stance reasoning. The experimental results show that these two parts are effective and complementary. For example, if we only use contrastive learning, the F1 score can reach 88.1%. And if we only use implicit stance reasoning, the F1 score can reach 87.3%. Only when using both of them can the F1 score reach 91.7%.

It is interesting that CL-ISR also has good generalization ability on cross-domain experiments. For example, if we transfer from FakeNewsNet to PHEME, the F1 score can reach 84.1%, which is significantly better than the baseline models, such as BERT and RoBERTa. It is known that generalized representation learned by contrastive learning can weaken the impact of domain dependence, which makes CL-ISR can be applied in different data distributions and enhance its practical application.

In summary, we propose a new and effective method CL-ISR for misleading social media content detection. Contrastive learning and implicit stance reasoning enhance the model's ability to understand semantic information and extract misleading strategies. Related cross-domain work in machine vision also highlights the potential of deep learning for complex detection tasks [16]. In the future, we will continue to improve the model structure, and try to apply it in other application contexts and more diverse data sources, to further improve the performance of the model. We hope the research can provide technical support for combating misinformation on social media and create a safer and healthier social media environment.